\pdfoutput=1

\documentclass[11pt]{article}
\usepackage[final]{acl}
\usepackage{times}
\usepackage{latexsym}
\usepackage{graphicx}
\usepackage{booktabs}
\usepackage{multirow}
\usepackage{amsmath}
\usepackage{xcolor}
\usepackage{url}
\usepackage{enumitem}
\usepackage{tabularx}
\usepackage{array}
\usepackage{booktabs}
\usepackage{colortbl}  
\usepackage{xcolor}
\usepackage{mdframed}
\usepackage{placeins}

\newcommand{\posdelta}[1]{\textcolor{teal}{#1}}
\newcommand{\negdelta}[1]{\textcolor{red!70!black}{#1}}

\usepackage[most]{tcolorbox}
\usepackage{xcolor}
\usepackage{soulutf8}
\usepackage{enumitem}
\usepackage{xurl}      
\definecolor{prov}{HTML}{2563EB}
\definecolor{treaty}{HTML}{059669}
\definecolor{resol}{HTML}{D97706}
\definecolor{casec}{HTML}{DC2626}
\definecolor{court}{HTML}{7C3AED}
\definecolor{soft}{HTML}{DB2777}
\definecolor{domlaw}{HTML}{0891B2}

\newcommand{\ent}[3]{%
  \begingroup
  \sethlcolor{#2!18}%
  \hl{#1}\,\textcolor{#2!85!black}{\scriptsize\textsf{#3}}%
  \endgroup
}

\newcommand{\PROV}[1]{\ent{#1}{prov}{PROV}}
\newcommand{\TRTY}[1]{\ent{#1}{treaty}{TRTY}}
\newcommand{\RESL}[1]{\ent{#1}{resol}{RESL}}
\newcommand{\CASE}[1]{\ent{#1}{casec}{CASE}}
\newcommand{\CRTI}[1]{\ent{#1}{court}{CRTI}}
\newcommand{\SOFT}[1]{\ent{#1}{soft}{SOFT}}
\newcommand{\DLAW}[1]{\ent{#1}{domlaw}{DLAW}}

\newtcolorbox{nerexample}[1]{
  breakable,
  colback=gray!2,
  colframe=gray!35,
  boxrule=0.3pt,
  arc=1.5pt,
  left=4pt,
  right=4pt,
  top=3pt,
  bottom=3pt,
  fonttitle=\bfseries\footnotesize,
  title=#1
}

\title{IntLawNER: A Named Entity Recognition Dataset and Benchmark in International Law}

\author{
  Genis Skura \and Roland Bouffanais \and Didier Wernli \\
  Department of Computer Science, Faculty of Science \\
  Global Studies Institute \\
  University of Geneva, Switzerland
}

\begin{document}
\maketitle

\begin{abstract}
International law provides the normative framework through which states coordinate action, regulate armed conflict, and protect 
human rights, yet its texts remain without 
token-level named entity recognition (NER) resources. We introduce \noindent\textbf{IntLawNER}\footnote{Dataset: \url{https://huggingface.co/datasets/Geo14/IntLawNER}\\
Code: \url{https://github.com/mrgeooo14/IntLawNER}}, a NER dataset and benchmark for codified sources of international law, covering 2,987 gold-annotated sentences and 8,094 entity spans from International Court of Justice (ICJ) decisions, UN Security Council resolutions, and European Court of Human Rights (ECtHR) judgments, annotated with seven institution-specific entity types. We construct IntLawNER with a cost-effective hybrid algorithmic–agentic pipeline that reduces 468k source sentences to a compact annotation set through candidate retrieval, LLM-based vetting, and human review, with 89.6\% of gold spans accepted unchanged from the silver layer. However, the silver-to-gold analysis reveals that human--machine aggregate agreement metrics can be misleading in domain-specific NER: Cohen's $\kappa$=0.964 on boundary-matched spans masks a macro-F1 of 0.753 when missing entities, boundary errors, and label corrections are included. Benchmark shows that zero-shot span-based GLiNER collapses on entity types dependent on institutional function rather than surface form
 (0.243 micro-F1), while fine-tuned transformers struggle on rare labels. Carefully selected few-shot examples that demonstrate label contrasts improve every LLM over zero-shot prompting, with Claude Opus 4.6 reaching the best score of 0.873 micro-F1. We release 
\textsc{IntLawNER} as a benchmark and reusable resource for extracting references in international legal texts.
\end{abstract}

\section{Introduction}
\label{sec:intro}

International law provides the normative framework through which states coordinate action, regulate armed conflict, protect human rights, and resolve disputes. These functions are embodied in entity-diverse treaties, judicial decisions, resolutions, and other legal instruments, yet no NER benchmark targets them.

Existing resources for NER remain concentrated in domestic law as benchmarks now exist for German court decisions \cite{leitner2019fine,leitner2020dataset}, Indian court judgments \cite{kalamkar2022ner}, Greek legislation \cite{angelidis2018ner}, Turkish legal texts \cite{turkish_ner}, English contracts \cite{deeplearning2024contracts}, and general legal text \cite{karamitsos2025legner}.
These resources have catalyzed work on legal language models \cite{chalkidis2020legalbert} and multi-task NLU benchmarks \cite{chalkidis2022lexglue}. Nevertheless, none of these benchmarks cover \emph{international legal texts}.

International legal text differs from domestic legal text in ways that matter for NER, which we illustrate with one example per corpus. First, the same sentence can contain references to both domestic and international instruments. ECtHR judgments, for example, cite national criminal codes alongside the European Convention on Human Rights, and the model must classify each correctly despite identical syntactic frames. General-purpose pipelines collapse this distinction (e.g. spaCy assigns both the same LAW label).  UNSC resolutions contain binding treaty references, numbered resolution cross-references, and non-binding instruments (action plans, conferences, declarations) within entity-dense operative paragraphs, distinctions that models must reproduce based on legal status rather than surface form. ICJ case citations pose a further challenge: where a single long entity span such as \textit{"Oil Platforms (Islamic Republic of Iran v. United States of America), Preliminary Objection, Judgment, I.C.J. Reports 1996 (II), p. 812"} must be annotated as one contiguous IL\_CASE\_CITATION, even though it embeds what might superficially appear to be a treaty reference, a court decision type, and a reporter reference.  Such spans routinely exceed 30 tokens, well beyond the effective window of span-enumeration models like GLiNER.

We aim to make four contributions to the emerging Legal NLP community:

\begin{enumerate}[leftmargin=*,topsep=2pt,itemsep=1pt]
\item \textbf{IntLawNER}, a NER dataset for international legal texts, comprising 2{,}987 gold-annotated sentences across three institutional corpora (ICJ, UNSC, ECtHR) with a 7-type entity schema targeting named and 
codified legal authorities (\S\ref{sec:data}).

\item \textbf{A hybrid algorithmic--agentic NER dataset construction pipeline} for selecting $n$ high-quality annotation candidates from 468k source sentences. The algorithmic pipeline first selects diverse sentence-level candidates using weak labeling, Legal-BERT embeddings, spherical $k$-means clustering, and diversity-aware scoring. Next, an LLM-based agentic vetting operation narrows the selection based on corpus coverage monitoring, candidate selection, linguistic diversity control, and quality auditing (\S\ref{sec:methodology}).

\item \textbf{A controlled empirical analysis of the silver-to-gold gap}
in domain-specific NER: 89.6\% of gold spans match the LLM silver-labels, yet this agreement falls to macro-F1=0.753 once boundary
adjustments and missing entities are scored, with errors concentrated in a
systematic confusion between judicial decisions and standing court instruments
(\S\ref{sec:experiments}).

\item \textbf{An empirical NER benchmark} comparing rule-based extraction, fine-tuned transformers, zero-shot span-based models, and LLMs on zero-shot and few-shot. The integration of ECtHR language is particularly informative, given their inclusion in Legal-BERT pre-training \citep{chalkidis2020legalbert}, enabling us to test whether its representations generalize to international legal entity types for which they were not explicitly trained. The results reveal a clear hierarchy: regex and GLiNER underperform (0.345 and 0.243 micro-F1), fine-tuned XLM-RoBERTa reaches 0.849 micro-F1 while the best few-shot LLM (Claude Opus 4.6) reaches 0.873.

\end{enumerate}
\section{Related Work}
\label{sec:related}

\paragraph{Legal NER Benchmarks.}
Legal NER has grown substantially but remains concentrated in domestic law. For example, 
\citet{leitner2019fine,leitner2020dataset} created a 67k-sentence German legal NER dataset with 19 entity types from federal court decisions.
\citet{kalamkar2022ner} developed a 14-type schema for Indian court judgments, with jurisdiction-specific entities (judge, petitioner, statute, provision).
\citet{karamitsos2025legner} fine-tuned domain-adapted BERT for legal NER and text anonymization.
Contract NER benchmarks achieve F1=0.94 with Legal-BERT \cite{deeplearning2024contracts}.
Before designing our schema we reviewed these inventories: \citet{leitner2019fine} distinguish law, ordinance, European legal norm and court decision; \citet{kalamkar2022ner} contract and anonymization schemas \citep{deeplearning2024contracts,karamitsos2025legner} target parties and clauses. None has a type for treaties, IO resolutions, soft-law instruments, or the standing procedural instruments of courts. We therefore reuse the three transferable types (statute, provision, precedent) under aligned names and add four.

\paragraph{Legal Language Models and Benchmarks.}
\citet{chalkidis2020legalbert} pre-trained BERT on 12GB of EU legislation, ECtHR cases, UK legislation, and US court opinions, establishing Legal-BERT as the standard domain-adapted encoder for legal NLP.
LexGLUE \cite{chalkidis2022lexglue} provides a multi-task benchmark for legal language understanding, including ECtHR violation prediction, but includes no token-level NER task. Consequently, IntLawNER could be viewed as the NER complement to LexGLUE.

\paragraph{LLMs as NER Annotators.}
The use of LLMs as annotators for NER dataset construction is established as a scalable alternative to manual annotation.
GPT-NER \cite{wang2023gptner} reformulated sequence labeling as text generation.
UniversalNER \cite{zhou2024universalner} used ChatGPT annotations to train distilled open-domain NER models.
FiNERweb \cite{finerweb2024} scaled LLM-based NER pipelines to the web.
PromptNER \cite{ashok2024promptner} showed that detailed type descriptions improve zero-shot NER. IntLawNER provides a controlled measurement of the human-machine agreement for international legal NER, revealing that high aggregate agreement score ($\kappa$=0.964) might mask a catastrophic label-level failure (F1=0.044 for IL\_COURT\_INSTR).

\paragraph{Zero-Shot NER Models.}
GLiNER \cite{zaratiana2024gliner} introduced a compact bidirectional-transformer model for zero-shot NER using span-label similarity matching, achieving strong results on standard benchmarks.
\citet{li2023evaluating} evaluated ChatGPT on information extraction tasks. Our candidate selection strategy is related to recent self-improving NER methods that use LLM-generated annotations over unlabelled corpora and then filter them for reliability \citep{xie2024selfimproving}. Unlike this line of work, which relies on self-annotated labels at inference time, we pass the LLM's selected candidates to human adjudication, producing a manually audited gold standard for international legal NER, rather than general English.

\paragraph{Agentic Pipelines for Data Curation.}
\citet{sumers2024cognitive} surveyed cognitive architectures for language agents.
A multi-agent framework for low-resource NER \cite{multiagent2025ner} uses knowledge retrieval and reflective analysis at inference time while \citet{kim2025lpdata} validated purpose-driven dataset pipelines.
IntLawNER's Phase 2 pipeline operates at dataset \emph{construction} time rather than inference time: a four-node LangGraph agentic network (coverage monitor, selector, diversity gate, QA) iteratively selects and annotates candidates, producing silver labels that a human annotator then verifies.

\section{Data and Task Definition}
\label{sec:data}

\subsection{Source Corpora}

IntLawNER draws sentences from three public institutional corpora representing distinct venues of international law (Table~\ref{tab:corpora}).

\begin{table}[t]
\centering
\small
\begin{tabular}{@{}lrr@{}}
\toprule
\textbf{Corpus} & \textbf{Document Count} & \textbf{Character Count} \\
\midrule
CD-ICJ & 2{,}289 & 90{,}300{,}190 \\
ECtHR-PCR & 11{,}000 & 111{,}016{,}058 \\
UNSC & 2{,}816 & 21{,}210{,}837 \\
\midrule
\textbf{Total} & \textbf{16{,}105} & \textbf{222{,}527{,}085} \\
\bottomrule
\end{tabular}
\caption{Source corpora. CD-ICJ: ICJ judgments, orders, advisory opinions, and separate/dissenting opinions \citet{fobbe2023}. UNSC: Security Council resolutions 1946-2026. ECtHR: European Court of Human Rights case judgments \citet{ecthr_2024}.} \label{tab:corpora}
\end{table}

The CD-ICJ is derived from Fobbe's Corpus \cite{fobbe2023} while UNSC contains 2,816 resolutions (1946–2026), collected from the UN Digital Library. The ECtHR component is derived from ECtHR-PCR, a dataset of European Court of Human Rights judgments \citet{ecthr_2024}, from which we extract individual sentences. All texts are in English.

\subsection{Entity Schema}

We define seven entity types targeting named legal instruments and authorities (Table~\ref{tab:label_examples}, Appendix~\ref{app:definitions}).

\begin{table}[t]
\centering
\small
\begin{tabularx}{\linewidth}{@{}l>{\raggedright\arraybackslash}X@{}}
\toprule
\textbf{Label} & \textbf{Entity Span Example} \\
\midrule
\texttt{IL\_TREATY} & \emph{the United Nations Convention on the Law of the Sea} \\
\texttt{IL\_PROVISION} & \emph{Article 2, paragraph 4} \\
\texttt{IL\_RESOLUTION} & \emph{Security Council resolution 1373 (2001)} \\
\texttt{IL\_COURT\_INSTR} & \emph{the Rules of Court of the ICJ} \\
\texttt{IL\_CASE\_CITATION} & \emph{Nicaragua v. United States of America} \\
\texttt{IL\_SOFT\_LAW} & \emph{the 2030 Agenda for Sustainable Development} \\
\texttt{IL\_DOMESTIC\_LAW} & \emph{the German Basic Law} \\
\bottomrule
\end{tabularx}
\caption{IntLawNER label inventory with representative examples from the gold annotations. See Appendix~\ref{app:definitions} for label definitions}
\label{tab:label_examples}
\end{table}

\textbf{IL\_RESOLUTION} and \textbf{IL\_PROVISION} are reliably identified from surface form: numbered resolution formats (\textit{resolution 1970 (2011)}, \textit{S/RES/2589}) and structural markers (\textit{Article}, \textit{Chapter}, \textit{§}) are high-precision cues. The main challenge for provisions is decomposition: \emph{Article 38, paragraph 5, of the Rules of Court} is decomposed into two entities, not one merged span.

\textbf{IL\_DOMESTIC\_LAW} covers national legislation and EU secondary legislation (regulations and directives adopted by EU institutions under Art. 288 TFEU), which are adopted through internal legislative procedures rather than negotiated as treaties and requires legal context reasoning from models.

\textbf{IL\_TREATY} and \textbf{IL\_SOFT\_LAW} require legal-status reasoning. Per VCLT Art.~2(1)(a) \citep{vclt1969}, a treaty is any international agreement concluded between states ``whatever its particular designation''---so an instrument titled \emph{Declaration} may be binding or non-binding depending on its legal effect. IL\_SOFT\_LAW captures the non-binding counterpart: declarations, action plans, guidelines, and frameworks adopted without inter-state obligations (e.g., the Universal Declaration of Human Rights). Both label types co-occur in identical syntactic contexts, making the distinction challenging without semantic reasoning.

Another difficult boundary exists between \textbf{IL\_COURT\_INSTR} and \textbf{IL\_CASE\_CITATION}. The first covers standing procedural instruments adopted by international judicial bodies under their own authority (\textit{Rules of Court}, \textit{Practice Directions}) while the second covers references to decisions in specific cases. Both share identical syntactic frames, so a model must distinguish persistent rules from case-specific decisions entity spans without lexical cues.

\subsection{Dataset Statistics}
\label{sec:dataset-statistics}

In our three-corpora dataset (Table~\ref{tab:corpora}), ECtHR dominates the source pool (11,000 of 16,105 documents; 68\%). Therefore, corpus-balance weights in Phase~1 scoring and corpus-cap agent directives were established to enforce a balanced gold set 38.5\%-35.0\%-26.5\% 
(ECtHR/CD-ICJ/UNSC). This was a deliberate choice, since each corpus contributes near-exclusive 
label mass: 94.9\% of all IL\_RESOLUTION spans originate in 
UNSC, 94.6\% of IL\_DOMESTIC\_LAW in ECtHR, and 75.6\% of 
IL\_CASE\_CITATION in CD-ICJ, in the final gold set (Figure~\ref{fig:entitydistribution}).

A model cannot learn all 
seven entity types from any single corpus. UNSC text is 
dominated by resolutions and treaty references but almost no case citations (4 spans) or domestic 
legislation (22 spans); ECtHR is rich in domestic laws but devoid of resolutions (7 spans); CD-ICJ carries the bulk of case 
law and court instruments but minimal domestic legislation (43 
spans) (Figure~\ref{fig:entitydistribution}). 

\section{Methodology}
\label{sec:methodology}

\begin{figure*}[t]
  \includegraphics[width=0.92\linewidth]{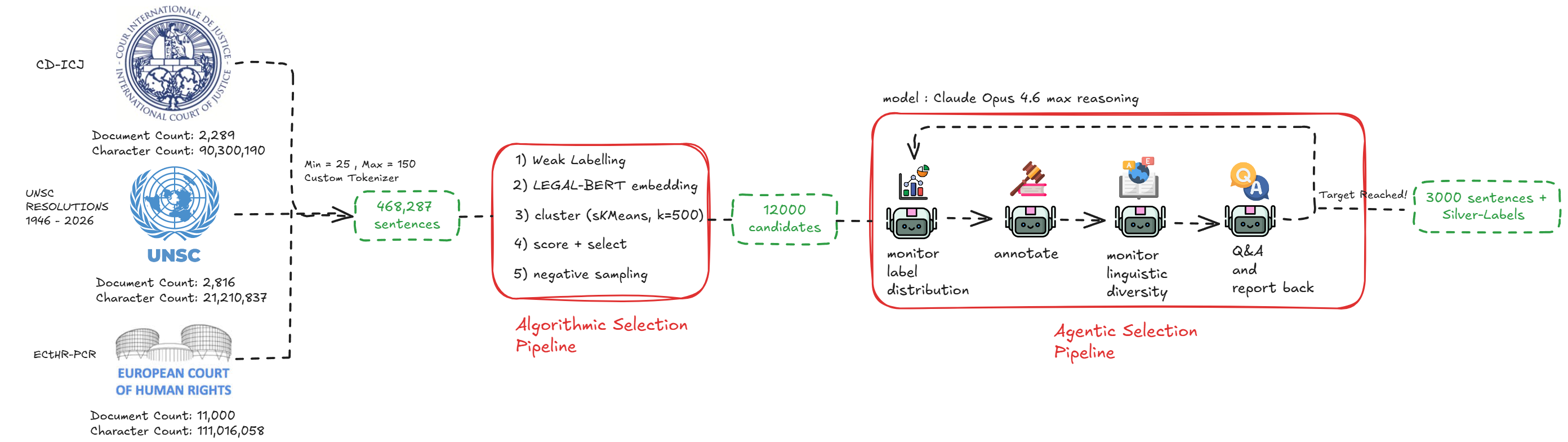}
  \hfill
    \caption{IntLawNER construction pipeline. Phase 1 uses algorithmic methods to select diverse, entity-rich candidates. Phase 2 applies a LangGraph-based agentic loop to select and annotate sentences.}
    \label{fig:pipeline}
\end{figure*}

The IntLawNER dataset is constructed in three phases: (1)~algorithmic candidate selection, (2)~agentic vetting (Fig.~\ref{fig:pipeline}), and (3)~gold annotation.

\subsection{Phase 1: Algorithmic Candidate Selection}

\paragraph{Sentence Tokenization.}
We process 16{,}105 documents using pySBD as a sentence
tokenizer, and post-processing for idiosyncratic legal boundaries (numbered
paragraphs, inline citations). Running headings, footnotes, tables, and page
numbers are stripped, while paragraph numbers are kept as part of
provision and case spans. Candidates are filtered by length (25--150
tokens) and quality checks (finite-verb, near-duplicate, and noise filters),
which also discard residual structural artifacts, yielding 468{,}287 valid
sentences.

\paragraph{Stage 1: Weak Labeling.}
Each sentence is labeled using two complementary methods.
First, spaCy's \textit{en\_core\_web\_trf} model provides LAW labels.
Second, regex patterns target seven IL\_ entity types using surface patterns (e.g., ``resolution \textbackslash d+'' for IL\_RESOLUTION, ``Article \textbackslash d+'' for IL\_PROVISION.

\paragraph{Stage 2: Sentence Embedding.}
Each sentence is encoded using Legal-BERT (\textit{legal-bert-base-uncased}) \citealt{chalkidis2020legalbert}) with mean pooling and L2 normalization, producing 768-dimensional embeddings.

\paragraph{Stage 3: Spherical $k$-Means Clustering.}
We apply spherical $k$-means clustering with $k$=500 using FAISS \cite{johnson2019billion}, to ensure that  selected candidates cover the topical diversity from each corpus.

\paragraph{Stage 4: Diversity-Aware Scoring and Selection.}
Each sentence receives a composite score:
\begin{equation}
s_i = d_i \cdot (1 + v_i) \cdot w_{c_i}
\end{equation}
where $d_i$ is entity density (number of detected entities normalized by sentence length), $v_i$ is label diversity (number of distinct entity types), and $w_{c_i}$ is a corpus balance weight ensuring proportional representation.

We select the top 10{,}000 positive candidates by composite score and 2{,}000 negative samples drawn at random from the detector-empty remainder, yielding 12{,}000 candidates. Negative samples supply empty sentences for evaluation, and serve as a recall channel for entities the weak labelers do not detect (Appendix~\ref{app:provenance}).

\subsection{Phase 2: Agentic Vetting}

Phase 2 implements a four-node iterative loop using LangGraph \cite{langgraph2024} to select a target of $n$ = 3{,}000 sentences from the 12{,}000 candidate pool while producing silver annotations. 
The agents were grounded in the same label glossary and annotation
guidelines used by human annotators, with definitions derived from four
institutional sources: the \textit{International Law Handbook},
the \textit{UN Security Council Handbook}, the \textit{International Court
of Justice Handbook}, and the \textit{European Convention on Human Rights}
\citep{un_international_law_handbook_2017,security_council_report_2019,icj_handbook_2019,echr_convention}. 

The full glossary is provided in Appendix~\ref{app:definitions}.

\paragraph{Coverage Monitor.}
An agent assumes the role of team leader and iteratively analyzes label and corpus distributions. It identifies underrepresented categories and issues corpus or label priority directives to downstream agents. (e.g., ``prioritize IL\_SOFT\_LAW from UNSC'').

\paragraph{Selector.}
An agent selects batches of sentences, produces the 7-label silver annotations based on the provided label glossary and annotation guidelines shared with the human annotators, and identifies hard negatives---sentences that contain entities of other types or none at all. Our labels are function-defined and require glossary legal reasoning beyond what surface-pattern or span-similarity methods can capture, hence the need for silver labeling through LLMs and this agent.

\paragraph{Diversity Gate.}
An agent filters redundant candidates by evaluating sentences across five diversity axes: syntactic position, linguistic diversity, entity decomposition complexity, surface-form alternation of the same entity, and cross-corpus variation. This approach enhances the Phase 1 diversity gate, compensating for the limitations of embedding-based clustering on linguistically similar sentences.

\paragraph{QA Gate.}
A periodic (every 5 iterations) quality audit agent samples annotated sentences, checks their consistency against the label glossary, and feeds the audit back to the coverage monitor.

Processing 12{,}000  candidates, the loop ran for 250 iterations ($\sim$800 LLM calls) until reaching the target of $n$ = 3{,}000 silver-annotated sentences.

The example of agentic reasoning during one iteration can be found in Appendix ~\ref{app:agentsinaction}. All agents are Claude Opus 4.6 models.

\subsection{Phase 3: Gold Human Annotation}

Silver labels were imported into Label Studio for human annotator review. Thirteen sentences are dropped during review, yielding a final gold set of 2,987 sentences with 8,094 spans. Of those, 7,250 (89.6\%) were accepted from the LLM without modification. The remaining 10.4\% break down 
into : 201 are label-only corrections, 289 are boundary-only adjustments, 29 change both label and boundary, and 325 are added spans with no overlapping silver span where the LLM missed entirely. A more detailed breakdown of these differences is discussed in section ~\ref{sec:silver-gold}.

\section{Experiments}
\label{sec:experiments}

\subsection{Evaluation Setup}

The final gold-labeled dataset contains 2{,}987 sentences with 8{,}094 entity spans (gold) [Fig~\ref{fig:entitydistribution}, Table~\ref{tab:splits}]. We use an approximately 70/10/20 train/dev/test split stratified by corpus, with document-level splitting to prevent data leakage. Approximately 14.5\% of sentences are negatives (no entities). Mean annotation density is 2.71 spans per sentence.

All systems are evaluated on the gold test set (605 sentences, 1{,}660 entity spans) using entity-level precision, recall, and F1 with strict BIO matching via seqeval \cite{seqeval}.
We report per-label F1 and micro/macro averages.

\paragraph{Systems.}
We evaluate four system categories:
(1)~A \textbf{regex baseline} using deterministic pattern matching for all 7 entity types.
(2)~\textbf{GLiNER} (\textit{urchade/gliner\_large-v2.1}; \citealt{zaratiana2024gliner}) with a $0.3$ threshold based on its best micro-F1 score on the development set (Table~\ref{tab:gliner_dev_threshold}).
(3)~\textbf{Nine LLMs} in zero-shot and 5-shot settings with the full label glossary: Claude Sonnet 4.6; Opus 4.6 and Opus 4.7 (Anthropic); GPT-5.5 (OpenAI); Gemini 2.5 Pro (Google); Llama 3.3 70B (Meta); DeepSeek V4 Pro (DeepSeek); Grok 4.3 (xAI); and Mistral Large 2512.
We tune the number of few-shot demonstrations on the development set (Table~\ref{tab:llm_k_sweep_dev}). Since k=5 is best for two of the strongest models and is effectively tied with k=3 on average, we select k=5 for the few-shot test-set setting of LLMs. For this few-shot prompting, we manually selected gold-annotated sentences that best illustrate difficult label boundaries and provide representative decision cues for the LLMs (Appendix~\ref{app:fewshot_examples}).
(4)~Two \textbf{fine-tuned transformers}: Legal-BERT \cite{chalkidis2020legalbert} and XLM-RoBERTa-large \cite{conneau2020xlmr}, trained for BIO token classification on the IntLawNER training set.
Fine-tuned models use learning rate 3e-5, batch size 8 with gradient accumulation 4, 8 epochs with early stopping (patience 2) which were selected as the best hyperparameters upon benchmarking on the development set. They are evaluated over 5 random seeds : \{14, 42, 123, 7, 99\} (Table~\ref{tab:transformer_seed_summary}).

\subsection{Main Results}

Table~\ref{tab:main} reports main results.

\begin{table*}[t]
\centering
\small
\setlength{\tabcolsep}{3.5pt}
\begin{tabular}{@{}llccccccccc@{}}
\toprule
& \textbf{System} & \textbf{PROV} & \textbf{TRTY} & \textbf{RESL} & \textbf{DLAW} & \textbf{CASE} & \textbf{SOFT} & \textbf{CRTI} & \textbf{Mi-F1} & \textbf{Ma-F1} \\
\midrule
\multicolumn{11}{@{}l}{\textit{Baselines}} \\
& Regex            & .504 & .222 & .425 & .097 & .155 & .034 & .710 & .345 & .307 \\
& GLiNER           & .322 & .065 & .617 & .093 & .023 & .000 & .048 & .243 & .167 \\
\midrule
\multicolumn{11}{@{}l}{\textit{LLMs --- zero-shot}} \\
& GPT-5.5          & .878 & .816 & .875 & .848 & .606 & .441 & .788 & .827 & .750 \\
& Gemini 2.5 Pro   & .881 & .810 & .893 & .794 & .596 & .522 & .706 & .822 & .743 \\
& DeepSeek V4 Pro  & .868 & .803 & .905 & .769 & .530 & .400 & .667 & .806 & .706 \\
& Opus 4.6         & .893 & .704 & .968 & .604 & .644 & .175 & .686 & .793 & .668 \\
& Mistral Large    & .906 & .716 & .913 & .685 & .470 & .256 & .703 & .784 & .664 \\
& Opus 4.7         & .872 & .658 & .963 & .581 & .635 & .214 & .588 & .767 & .644 \\
& Grok 4.3         & .886 & .604 & .890 & .555 & .533 & .367 & .545 & .739 & .626 \\
& Sonnet 4.6       & .839 & .550 & .957 & .544 & .558 & .098 & .647 & .723 & .599 \\
& Llama 3.3 70B    & .834 & .605 & .809 & .600 & .424 & .188 & .722 & .698 & .598 \\
\midrule
\multicolumn{11}{@{}l}{\textit{LLMs --- 5-shot}} \\
& Opus 4.6         & .925 & .840 & .959 & .852 & .614 & .708 & .743 & \textbf{.873} & \textbf{.806} \\
& Sonnet 4.6       & .925 & .812 & .961 & .843 & .697 & .600 & .743 & .871 & .797 \\
& GPT-5.5          & .934 & .810 & .951 & .843 & .622 & .550 & .727 & .861 & .777 \\
& Opus 4.7         & .902 & .803 & .968 & .808 & .576 & .514 & .686 & .842 & .751 \\
& Gemini 2.5 Pro   & .915 & .802 & .904 & .835 & .569 & .463 & .688 & .836 & .739 \\
& Mistral Large    & .909 & .790 & .934 & .765 & .545 & .432 & .632 & .824 & .715 \\
& DeepSeek V4 Pro  & .913 & .772 & .893 & .763 & .500 & .466 & .686 & .811 & .713 \\
& Grok 4.3         & .904 & .691 & .870 & .661 & .561 & .384 & .706 & .779 & .682 \\
& Llama 3.3 70B    & .874 & .673 & .813 & .682 & .577 & .329 & .684 & .757 & .662 \\
\midrule
\multicolumn{11}{@{}l}{\textit{Fine-tuned transformers (mean over 5 seeds)}} \\
& XLM-R-large      & .941 & .798 & .911 & .847 & .592 & .423 & .643 & .849 & .736 \\
& Legal-BERT       & .924 & .743 & .932 & .816 & .558 & .078 & .338 & .820 & .627 \\
\bottomrule
\end{tabular}
\caption{Per-label F1 on the IntLawNER test set. Mi-F1 and Ma-F1 denote
micro- and macro-averaged F1. For fine-tuned transformers, entries are
averaged over five random seeds (14, 42, 123, 7, 99). Systems are sorted by
Mi-F1 within each block. 95\% confidence intervals from 1{,}000 document-level
bootstrap resamples are reported in Table~\ref{tab:ci-micromacro} and per-label Table~\ref{tab:ci-perlabel}}.
\label{tab:main}
\vspace{2pt}
\begin{flushleft}
\footnotesize
\emph{Abbreviations:} PROV=provision, TRTY=treaty, RESL=resolution, DLAW=domestic law, CASE=case citation, SOFT=soft law, CRTI=court instrument.
\end{flushleft}
\end{table*}    

The best overall system is Claude Opus 4.6 with 5-shot prompting 
(micro-F1=0.873, macro-F1=0.806), followed by Sonnet 4.6 5-shot 
(0.871 micro, 0.797 macro) and GPT-5.5 5-shot 
(0.861 micro, 0.777 macro). XLM-RoBERTa-large 
achieves 0.849 micro-F1.

Three findings emerge from the main results.
First, \textbf{few-shot prompting improves micro-F1 across LLMs}.
Sonnet~4.6 improves from .723 to .871 micro-F1 (+14.8 points);
Opus~4.6 from .793 to .873 (+8.0); Opus~4.7 from .767 to
.842 (+7.4); and Mistral Large from .784 to .824 (+4.1).
The gains are especially visible on the minority labels that require reasoning rather than surface cues. For
IL\_SOFT\_LAW, Opus~4.6 improves from .175 to .708 (+53.3
points), Sonnet~4.6 from .098 to .600 (+50.2), and Opus~4.7
from .214 to .514 (+30.0). IL\_DOMESTIC\_LAW also improves
substantially for several models, most notably Sonnet~4.6
(.544 $\rightarrow$ .843, +29.9 points). (Table~\ref{tab:ablation})

Second, \textbf{fine-tuned transformers do not consistently outperform LLMs despite access to training data}. 
XLM-RoBERTa-large’s
seed-mean micro-F1 (0.849) sits at the lower edge of the confidence intervals of Opus 4.6 5-shot (0.873) and Sonnet 4.6 5-shot (0.871) while
Legal-BERT (0.820) is lower still; both, however, underperform on rare labels : IL\_SOFT\_LAW
and IL\_COURT\_INSTR, in comparison to few-shot LLMs without access to training data. The fine-tuned models reflect their weakness on minority classes with rare labels and labels that require international legal reasoning rather than surface cue-matching.

Third, \textbf{the newer Opus version does not improve domain-specific NER.}
Opus~4.7 scores below Opus~4.6 in both settings (5-shot 0.842 vs.\ 0.873), but
the intervals overlap, so we report this as the absence of an improvement rather
than a regression.

\paragraph{Error Concentration.}
Averaged over the LLM systems, performance is strongest on
IL\_RESOLUTION (mean F1=.913) and IL\_PROVISION (.892), followed by
IL\_TREATY (.737) and IL\_DOMESTIC\_LAW (.724). The main residual errors
are concentrated in the boundary- and function-sensitive labels:
IL\_CASE\_CITATION (.570), IL\_SOFT\_LAW (.395), and IL\_COURT\_INSTR
(.686), given their overlap
with case-citation language and treaties.

\subsection{Silver-to-Gold Gap Analysis}
\label{sec:silver-gold}

Table~\ref{tab:silver_gold} presents the silver-to-gold agreement 
analysis, comparing LLM-generated silver annotations against gold labels using exact
span matching (identical start, end, and label). This evaluation protocol captures label errors, boundary mismatches, and missing entities, making it stricter.

A natural concern is that anchoring candidate selection on weak labels
(spaCy~+~regex) biases the benchmark toward easily detected, surface-cued
references. A provenance audit (Appendix~\ref{app:provenance}) was conducted and shows: 57.3\% of
gold spans carry a label and surface form that Phase~1 never emitted across the entire 468k corpus, rising to 89--98\% on the function-defined labels (\textsc{case}, \textsc{soft},
\textsc{crti}).

\subsection{Zero-Shot vs.\ Few-Shot}

Table~\ref{tab:ablation} reports the effect of few-shot examples across all LLMs.

\begin{table}[t]
\centering
\small
\begin{tabular}{@{}lccc@{}}
\toprule
\textbf{Model} & \textbf{0-shot} & \textbf{5-shot} & \textbf{$\Delta$} \\
\midrule
Sonnet 4.6       & .723 & .871 & \posdelta{+.148} \\
Opus 4.6         & .793 & \textbf{.873} & \posdelta{+.080} \\
Opus 4.7         & .767 & .842 & \posdelta{+.075} \\
Llama 3.3 70B    & .698 & .757 & \posdelta{+.059} \\
Mistral Large    & .784 & .824 & \posdelta{+.040} \\
Grok 4.3         & .739 & .779 & \posdelta{+.040} \\
GPT-5.5          & .827 & .861 & \posdelta{+.034} \\
Gemini 2.5 Pro   & .822 & .836 & \posdelta{+.014} \\
DeepSeek V4 Pro  & .806 & .811 & \posdelta{+.005} \\
\bottomrule
\end{tabular}
\caption{Zero-shot vs.\ 5-shot micro-F1, sorted by gain 
($\Delta$). Sonnet~4.6 shows the largest improvement (+14.8 
points), rising from near-worst zero-shot to second-best 
5-shot.}
\label{tab:ablation}
\end{table}

All models improve with 5-shot 
examples. The gain ranges from +0.5 (DeepSeek~V4~Pro) to 
+14.8 (Sonnet~4.6) micro-F1 points, suggesting that boundary 
examples resolve a zero-shot calibration failure rather than 
compensating for missing capability. The improvement is 
disproportionately driven by minority labels: IL\_SOFT\_LAW 
shows the largest per-label gain across most models, consistent 
with the finding by \citet{min2022rethinking} that 
demonstrations teach task format rather than label correctness. 

\section{Discussion}
\label{sec:discussion}

\paragraph{$\kappa$ is misleading}
Our $\kappa$=0.964 on boundary-matched spans would pass any 
quality threshold. Full-span macro-F1 is 0.753, revealing the models' confusion on domain-specific complex entity spans. Reporting only 
aggregate agreement risk might mask label-level catastrophes 
invisible to $\kappa$. Hence, we recommend reporting per-label confusion 
matrices as a minimum standard for LLM-annotated NER datasets.

\paragraph{The minority labels}
The results reveal a clear capability ordering. GLiNER 
(micro-F1=0.243) collapses because its zero-shot label conditioning cannot access the full annotation guideline and label glossary. Its fixed span window 
truncates the 30+ token composite citations routine in ICJ text. 
Fine-tuned transformers perform substantially better but remain sensitive to rare labels: Legal-BERT drops to F1=0.338 on IL\_COURT\_INSTR (58 training spans) and 0.078 on IL\_SOFT\_LAW (140 spans). We intentionally preserve these categories at their natural frequency, since the dataset is designed to reflect distinctions in international legal texts rather than to adhere to machine learning.

\paragraph{Algorithmic Phase 1} Phase 1 is a down-sampler, not a quality filter: it reduces the 468k-sentence corpus to a cost-effective  pool for the agentic loop while preserving recall through corpus-balancing weights and negative sampling. A purely agentic benchmark would not require this, and a production system may ultimately be LLM-exclusive. However, we retain rule-based and BERT models to benchmark all three paradigms on this specialized domain while keeping the dataset governed by human adjudication.

\paragraph{Few-shot examples teach boundaries to LLMs}
Sonnet~4.6 gains +14.8 micro-F1 points with 5 examples. The 
improvement concentrates in minority labels and boundary 
decisions, consistent with \citet{min2022rethinking}. Our 
few-shot examples are designed to show (Appendix~\ref{app:fewshot_examples}). In principle, few-shot should demonstrate contrasts, not exemplars.

There is no improvement of Opus~4.7 over Opus~4.6 in both zero-shot (.767 
vs.\ .793) and 5-shot (.842 vs.\ .873). In our case, general capability 
improvements do not automatically transfer to domain-specific 
entity recognition, reinforcing the need for domain-level 
evaluation rather than reliance on aggregate benchmarks.

\paragraph{Cross-corpus performance.}
Given the skewed per-label entity distributions between our three corpora (\S\ref{sec:dataset-statistics}), cross-corpus generalization is a structural requirement of the performance benchmark. No model trained on a single corpus will encounter sufficient examples of all seven label types in the other two. We are listing this as a design property, not a limitation, since it tests models against three different registers of international legal text.

\paragraph{Pipeline generalizability.}
The two-phase architecture sentence-level dataset building pipeline should transfer to any domain where large unlabelled 
corpora exist and expert annotation is expensive, provided a glossary that guides the models. The key 
transferable component is the QA-to-Coverage feedback loop, 
which dynamically rebalances label and corpus distributions 
without manual intervention. Biomedical, financial, and 
clinical NER are plausible but untested targets for future work.

\section{Conclusion}
\label{sec:conclusion}

We introduced IntLawNER, a NER resource designed specifically for international legal codified references: 2,987 gold sentences across CD-ICJ, UNSC, and ECtHR, annotated with seven institutionally grounded entity types. The hybrid algorithmic--agentic dataset construction pipeline that combines algorithmic candidate retrieval, LLM-based selection, and a QA-to-Coverage feedback loop reduced 468k source sentences to a compact $n$=3000 annotation set while preserving corpus and label coverage. We hypothesize that the 89.6\% post-audit acceptance rate suggests that this predict-then-correct strategy might generalize to other high-cost NER domains with large unlabelled corpora.

The silver-to-gold analysis shows that high aggregate agreement can mask label-level failures: despite Cohen's $\kappa=0.964$ on boundary-matched spans, macro-F1 falls to 0.753, revealing model's confusion on domain-specific complex entity spans. 

The benchmark shows that IntLawNER is primarily a boundary-reasoning task: GLiNER collapses because its zero-shot label conditioning cannot encode the glossary, fine-tuned transformers struggle on generally rare labels, and every LLM improves its zero-shot score by learning institutionally defined label contrasts from few-shot examples, with the best one (Claude Opus 4.6) reaching 0.873 micro-F1.

Future work should subject the schema to validation from international law experts, then extend IntLawNER to additional languages; apply entity relation extraction (e.g., PROVISION--of--TREATY); and scale the schema to further international institutions, including WTO, ICC, and ITLOS materials. We release IntLawNER as a benchmark for evaluating codified legal authorities information extraction and as a reusable resource for building international legal knowledge networks that fuel retrieval-augmented generation systems in a domain where hallucinations are costly.

\section*{Limitations}

\paragraph{Schema validation and agreement.}
The $\kappa$ reported in \S\ref{sec:silver-gold} is human-vs-machine agreement, therefore it should not be interpreted as traditional IAA. The schema was designed and anchored in codified institutional sources (Appendix ~\ref{app:definitions}) rather than our subjective judgement or machine learning requirements, but it has not been independently reviewed by international-law experts.

\paragraph{Schema scope.}
The 7-type schema was derived from the entity distributions 
of three publicly available, NLP-ready corpora (CD-ICJ, UNSC, 
ECtHR) and does not claim to exhaustively cover international law. In particular, customary international law and general principles of law (Art. 38(1)(b)–(c) of the ICJ Statute) are typically expressed as doctrinal claims rather than discrete named entities, and cannot be reliably captured through entity extraction alone. IntLawNER therefore targets the visible, codified layer of international legal authority rather than the full spectrum of international law.

\paragraph{Dataset size and label sparsity.}
At $n$=2,987 gold sentences, IntLawNER is modest in scale. $n$ can be scaled as a parameter of the selection pipeline but the current value 
reflects the cost of expert gold annotation in a domain where 
qualified annotators, requiring both NLP training and 
international law knowledge, are scarce. 
IL\_COURT\_INSTR (83 total spans, 18 in test) and IL\_SOFT\_LAW 
(192 total, 37 in test) are particularly sparse; per-label F1 
on these categories should be interpreted with caution, as a 
single boundary error can shift scores by several points. We 
retained both at natural frequency rather than merging or 
upsampling, as their rarity is a genuine property of the domain.

\paragraph{Candidate pre-selection.}
Phase~1 bounds what the agents can see. We mitigate selection bias with
corpus-balanced sampling and detector-blind negative sampling, and the provenance
audit (Appendix~\ref{app:provenance}) indicates the gold set is not confined to
surface-cued entities; however, we do not directly compare Phase~1 against a solely agentic pipeline, that is
feeding corpus-balanced random sentences to the agents without scoring, which
would isolate its recall cost. We leave this comparison to future work.

\paragraph{Selection Pipeline Cost}
The agentic vetting pipeline (Phase 2) requires approximately 800 LLM calls at an estimated cost of \$140--\$240 depending on the model for our target $n$ = 3000.
This is substantially cheaper than manual annotation from scratch but may limit replication for under-resourced research groups. This excludes costs towards multiple-run evaluation of LLMs and their benchmarking for the best $k$ few-shot parameter.

\section*{Acknowledgments}
The work was funded with a grant from the Foundation for the University of Geneva. Additionally, we thank Mylan Evrard for the beneficial comments on the manuscript.

\paragraph{Data Usability}
We release the gold-annotated IntLawNER dataset with 
train/dev/test splits, few-shot prompt templates, and the 
full annotation guidelines, requiring no domain-specific 
preprocessing to integrate into existing NER pipelines. The annotation layer is released under CC-BY-4.0
(\url{https://huggingface.co/datasets/Geo14/IntLawNER}); the underlying legal texts
retain the terms of their original sources (ICJ, UNSC, ECtHR). The construction
pipeline, evaluation scripts, and per-system bootstrap intervals are available at
\url{https://github.com/mrgeooo14/IntLawNER}.

\paragraph{Ethics Statement}

IntLawNER is constructed from publicly available legal texts.
ICJ and UNSC documents are published by the United Nations under open access.
ECtHR judgments are publicly available through HUDOC. We acknowledge that NER systems for international law could be used for automated legal analysis; users should be aware that model outputs require expert verification and do not constitute legal advice.

\bibliography{intlawner}

\newpage


\appendix

\section{Label Distributions}
\label{app:appendixfigs}

\begin{figure}[!htbp]
  \includegraphics[width=1\linewidth]{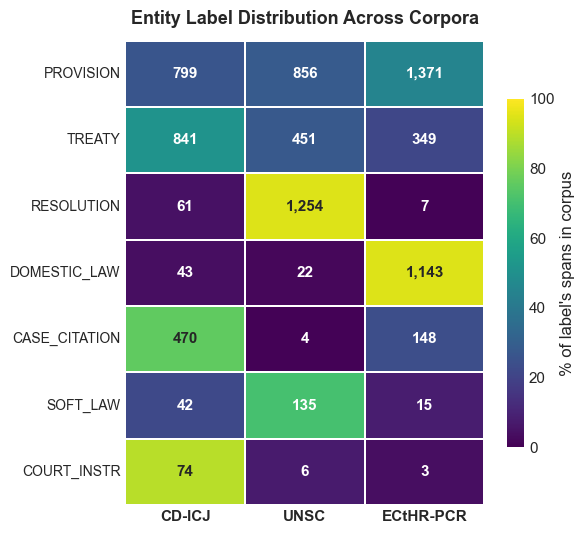}
  \hfill
    \caption{Distribution of the 7-label schema entities in the three corpora (final gold set after selection).}
    \label{fig:entitydistribution}
\end{figure}

\begin{table}[!htbp]
\centering
\small
\setlength{\tabcolsep}{2.5pt}
\begin{tabular}{@{}l rrr rrr r@{}}
\toprule
& \multicolumn{3}{c}{\cellcolor{gray!20}\textbf{Silver}} 
& \multicolumn{3}{c}{\cellcolor{yellow!20}\textbf{Gold}} 
& \\
\cmidrule(lr){2-4} \cmidrule(lr){5-7}
\textbf{Label} 
& \cellcolor{gray!20}\textbf{Total} 
& \cellcolor{gray!20}\textbf{\%} 
& \cellcolor{gray!20}
& \cellcolor{yellow!20}\textbf{Train} 
& \cellcolor{yellow!20}\textbf{Dev} 
& \cellcolor{yellow!20}\textbf{Test} 
& \textbf{$\Delta$} \\
\midrule
IL\_PROVISION     & 2{,}964 & 36.6 && 2{,}124 & 281 & 621 & \posdelta{+62} \\
IL\_TREATY        & 1{,}698 & 21.0 && 1{,}159 & 153 & 329 & \negdelta{-57} \\
IL\_RESOLUTION    & 1{,}299 & 16.0 && 874 & 152 & 296   & \posdelta{+23} \\
IL\_DOMESTIC\_LAW & 1{,}182 & 14.6 && 866 & 108 & 234   & \posdelta{+26} \\
IL\_CASE\_CITATION& 494     & 6.1  && 441 & 56  & 125   & \posdelta{+128} \\
IL\_SOFT\_LAW     & 181     & 2.2  && 140 & 15  & 37    & \posdelta{+11} \\
IL\_COURT\_INSTR  & 282     & 3.5  && 58  & 7   & 18    & \negdelta{-199} \\
\midrule
\textbf{Total spans} & 8{,}100 & 100 && 5{,}662 & 772 & 1{,}660 & $-$6 \\
\textbf{Sentences}   & 3{,}000 &     && 2{,}110 & 272 & 605     & --- \\
\bottomrule
\end{tabular}
\caption{Silver-to-gold label inventory. Silver: LLM-generated annotations from Phase~2. Gold: labels after audit (\S\ref{sec:silver-gold}). $\Delta$: gold $-$ silver span count.}
\label{tab:splits}
\end{table}

\begin{table}[t]
\centering
\small
\begin{tabular}{@{}lccc@{}}
\toprule
\textbf{Label} & \textbf{P} & \textbf{R} & \textbf{F1} \\
\midrule
IL\_PROVISION     & .953 & .932 & .942 \\
IL\_TREATY        & .898 & .923 & .910 \\
IL\_RESOLUTION    & .995 & .977 & .986 \\
IL\_DOMESTIC\_LAW & .964 & .943 & .954 \\
IL\_CASE\_CITATION& .651 & .510 & .572 \\
IL\_SOFT\_LAW     & .890 & .839 & .863 \\
IL\_COURT\_INSTR  & .029 & .096 & .044 \\
\midrule
Overall $\kappa$ & \multicolumn{3}{c}{0.964} \\
Macro F1         & \multicolumn{3}{c}{0.753} \\
\bottomrule
\end{tabular}
\caption{Silver-to-gold agreement by label. Silver labels are 
treated as predictions, gold labels as reference.}
\label{tab:silver_gold}
\end{table}

\begin{table}[t]
\centering
\small
\begin{tabular}{lrrrr}
\toprule
\textbf{Threshold} & \textbf{Precision} & \textbf{Recall} & \textbf{Micro-F1} & \textbf{Macro-F1} \\
\midrule
0.0 & 0.226 & 0.231 & 0.229 & 0.167 \\
0.1 & 0.343 & 0.229 & 0.275 & 0.190 \\
0.2 & 0.362 & 0.225 & 0.277 & 0.193 \\
\textbf{0.3} & \textbf{0.379} & \textbf{0.221} & \textbf{0.279} & \textbf{0.195} \\
0.4 & 0.391 & 0.211 & 0.274 & 0.194 \\
0.5 & 0.400 & 0.202 & 0.269 & 0.181 \\
\bottomrule
\end{tabular}
\caption{GLiNER threshold sweep on the development set. We select threshold $0.3$, which obtains the highest micro-F1. Macro-F1 is computed as the unweighted mean of label-level F1 scores.}
\label{tab:gliner_dev_threshold}
\end{table}

\begin{table}[t]
\centering
\small
\begin{tabular}{lrrr}
\toprule
\textbf{Model} & \textbf{Mean} & \textbf{Std.} & \textbf{95\% CI} \\
\midrule
Legal-BERT        & 0.820 & 0.011 & [0.807, 0.834] \\
XLM-RoBERTa-large & 0.849 & 0.002 & [0.847, 0.851] \\
\bottomrule
\end{tabular}
\caption{Fine-tuned transformer performance over five random seeds
(14, 42, 123, 7, and 99). Scores are test-set micro-F1. We report the mean,
standard deviation, and 95\% confidence interval over seeds, rather than
selecting the best test seed.}
\label{tab:transformer_seed_summary}
\end{table}

\begin{table}[t]
\centering
\small
\begin{tabular}{lrrrr}
\toprule
\textbf{Model} & \textbf{0-shot} & \textbf{$k=1$} & \textbf{$k=3$} & \textbf{$k=5$} \\
\midrule
Claude Sonnet 4.6 & 0.727 & 0.839 & 0.870 & 0.851 \\
Claude Opus 4.6   & 0.818 & 0.884 & 0.890 & 0.895 \\
GPT-5.5           & 0.839 & 0.856 & 0.866 & 0.879 \\
Llama-3.3-70B     & 0.669 & 0.764 & 0.758 & 0.759 \\
\midrule
Average           & 0.763 & 0.836 & 0.846 & 0.846 \\
\bottomrule
\end{tabular}
\caption{Development-set sweep over the number of demonstrations for few-shot LLM inference. Scores are micro-F1. We report zero-shot and $k=5$ on the test set: $k=5$ is the best setting for two of the strongest models and is effectively tied with $k=3$ on average, while providing a single fixed few-shot configuration across models.}
\label{tab:llm_k_sweep_dev}
\end{table}

\section{Provenance of Gold Spans}
\label{app:provenance}

To verify that IntLawNER is not merely a transcription of the Phase-1 weak
labels, we trace every gold span back to the Phase-1 detections and the Phase-2 silver labels. We call a span \emph{surface-unseen} if Phase~1 never tagged that exact string with that
label anywhere in the 468k-sentence corpus, and \emph{changed vs.\ silver} if it does not
exactly match a silver span. Table~\ref{tab:provenance} reports the results.
Across the gold set, 57.3\% of spans are surface-unseen, rising to 89--98\% on
the three function-defined labels (\textsc{case}, \textsc{soft}, \textsc{crti}),
precisely the categories where surface cues are least reliable.
The two surface-cued labels (\textsc{resl}, \textsc{prov}) are the only ones
where a majority of spans were surfaced by Phase~1, as expected.

\begin{table}[t]
\centering
\small
\setlength{\tabcolsep}{5pt}
\begin{tabular}{lrrr}
\toprule
Label & Gold & \multicolumn{1}{c}{Added/chg.} & \multicolumn{1}{c}{Surface} \\
      &      & \multicolumn{1}{c}{vs.\ silver} & \multicolumn{1}{c}{unseen} \\
\midrule
\textsc{prov} & 3{,}026 & 207   & 1{,}071  \\
\textsc{resl} & 1{,}322 & 30     & 413      \\
\textsc{trty} & 1{,}641 & 127    & 1{,}295 \\
\textsc{dlaw} & 1{,}208 & 69    & 1{,}044  \\
\textsc{case} & 622     & 305   & 554     \\
\textsc{soft} & 192     & 31   & 183     \\
\textsc{crti} & 83      & 75   & 81      \\
\midrule
All           & 8{,}094 & 844 (10.4\%)  & 4{,}641 (57.3\%) \\
\bottomrule
\end{tabular}
\caption{Gold-span provenance against Phase-1 weak labels.
\emph{Added/chg.\ vs.\ silver}: no exact $(\text{start}, \text{end}, \text{label})$
match in the silver layer.
\emph{Surface unseen}: the gold surface string was never emitted by Phase~1 with
that label anywhere in the corpus.
The blind-negative pool (detector-empty sentences) contributes a further 246
gold spans, confirming recovery of entities invisible to Phase~1.}
\label{tab:provenance}
\end{table}

\section{Confidence Intervals}
\label{app:ci}

We quantify test-set sampling uncertainty with a document-level bootstrap.
We resample the 434 test documents with replacement (1{,}000 iterations,
percentile method, seed 20260630), recompute every metric on each resample, and
report 95\% confidence intervals as the $[2.5, 97.5]$ percentiles.
Table~\ref{tab:ci-micromacro} gives micro/macro intervals for all systems;
Table~\ref{tab:ci-perlabel} gives per-label intervals.
Rare labels (\textsc{soft}, 37 test spans; \textsc{crti}, 18) carry intervals
spanning 30--40 F1 points, so per-label comparisons on these categories are
not informative.

\begin{table}[t]
\centering
\small
\setlength{\tabcolsep}{4pt}
\begin{tabular}{@{}lcc@{}}
\toprule
\textbf{System} & \textbf{Mi-F1 [95\% CI]} & \textbf{Ma-F1 [95\% CI]} \\
\midrule
\multicolumn{3}{@{}l}{\textit{Baselines}} \\
Regex   & .345 [.316,.373] & .307 [.266,.340] \\
GLiNER  & .243 [.218,.268] & .167 [.148,.186] \\
\midrule
\multicolumn{3}{@{}l}{\textit{LLMs --- zero-shot}} \\
GPT-5.5         & .827 [.804,.848] & .750 [.706,.787] \\
Gemini 2.5 Pro  & .822 [.798,.843] & .743 [.696,.782] \\
DeepSeek V4 Pro & .806 [.782,.827] & .706 [.657,.749] \\
Opus 4.6        & .793 [.764,.820] & .668 [.622,.704] \\
Mistral Large   & .784 [.757,.809] & .664 [.619,.704] \\
Opus 4.7        & .767 [.739,.794] & .644 [.593,.689] \\
Grok 4.3        & .739 [.710,.766] & .626 [.578,.668] \\
Sonnet 4.6      & .723 [.688,.754] & .599 [.552,.636] \\
Llama 3.3 70B   & .698 [.666,.726] & .598 [.552,.634] \\
\midrule
\multicolumn{3}{@{}l}{\textit{LLMs --- 5-shot}} \\
Opus 4.6        & .873 [.850,.893] & .806 [.764,.839] \\
Sonnet 4.6      & .871 [.850,.889] & .797 [.754,.832] \\
GPT-5.5         & .861 [.841,.879] & .777 [.732,.811] \\
Opus 4.7        & .842 [.819,.862] & .751 [.703,.789] \\
Gemini 2.5 Pro  & .836 [.812,.857] & .739 [.690,.776] \\
Mistral Large   & .824 [.800,.846] & .715 [.663,.758] \\
DeepSeek V4 Pro & .811 [.786,.833] & .713 [.666,.749] \\
Grok 4.3        & .779 [.749,.804] & .682 [.634,.722] \\
Llama 3.3 70B   & .757 [.729,.784] & .662 [.616,.703] \\
\midrule
\multicolumn{3}{@{}l}{\textit{Fine-tuned transformers}} \\
XLM-R-large      & .851 [.827,.872] & .734 [.690,.773] \\
Legal-BERT       & .801 [.771,.827] & .543 [.523,.562] \\
\bottomrule
\end{tabular}
\caption{Micro/macro F1 with 95\% document-level bootstrap confidence
intervals. The leading 5-shot systems (Opus~4.6, Sonnet~4.6, GPT-5.5) have
mutually overlapping micro-F1 intervals. Transformer intervals are bootstrapped over a single representative seed (123); seed-level variance is in Table ~\ref{tab:transformer_seed_summary}.}
\label{tab:ci-micromacro}
\end{table}

\newpage

\begin{table*}[t]
\centering
\scriptsize
\setlength{\tabcolsep}{3pt}
\begin{tabular}{@{}lccccccc@{}}
\toprule
\textbf{System} & \textbf{PROV} & \textbf{TRTY} & \textbf{RESL} & \textbf{DLAW} & \textbf{CASE} & \textbf{SOFT} & \textbf{CRTI} \\
\midrule
\multicolumn{8}{@{}l}{\textit{Baselines}} \\
Regex   & .504 [.456,.553] & .222 [.168,.281] & .425 [.353,.496] & .097 [.050,.153] & .155 [.080,.241] & .034 [.000,.116] & .710 [.470,.875] \\
GLiNER  & .322 [.275,.362] & .065 [.033,.099] & .617 [.546,.687] & .093 [.059,.133] & .023 [.000,.053] & .000 [.000,.000] & .048 [.000,.123] \\
\midrule
\multicolumn{8}{@{}l}{\textit{LLMs --- zero-shot}} \\
GPT-5.5         & .878 [.852,.903] & .816 [.776,.854] & .875 [.813,.929] & .848 [.796,.898] & .606 [.511,.696] & .441 [.281,.590] & .788 [.571,.945] \\
Gemini 2.5 Pro  & .881 [.855,.909] & .810 [.766,.852] & .893 [.835,.939] & .794 [.736,.850] & .596 [.504,.686] & .522 [.357,.674] & .706 [.444,.895] \\
DeepSeek V4 Pro & .868 [.835,.897] & .803 [.756,.843] & .905 [.866,.938] & .769 [.707,.829] & .530 [.426,.621] & .400 [.226,.595] & .667 [.400,.870] \\
Opus 4.6        & .893 [.864,.920] & .704 [.651,.757] & .968 [.943,.988] & .604 [.532,.674] & .644 [.528,.738] & .175 [.029,.353] & .686 [.429,.878] \\
Mistral Large   & .906 [.877,.933] & .716 [.664,.769] & .913 [.875,.942] & .685 [.619,.748] & .470 [.369,.569] & .256 [.132,.406] & .703 [.476,.867] \\
Opus 4.7        & .872 [.841,.899] & .658 [.603,.717] & .963 [.934,.984] & .581 [.515,.656] & .635 [.541,.723] & .214 [.048,.392] & .588 [.296,.818] \\
Grok 4.3        & .886 [.859,.913] & .604 [.544,.668] & .890 [.841,.930] & .555 [.480,.634] & .533 [.435,.638] & .367 [.203,.526] & .545 [.296,.759] \\
Sonnet 4.6      & .839 [.801,.875] & .550 [.487,.620] & .957 [.924,.981] & .544 [.466,.620] & .558 [.447,.661] & .098 [.000,.222] & .647 [.389,.857] \\
Llama 3.3 70B   & .834 [.799,.870] & .605 [.548,.663] & .809 [.743,.867] & .600 [.533,.670] & .424 [.323,.526] & .188 [.069,.327] & .722 [.483,.875] \\
\midrule
\multicolumn{8}{@{}l}{\textit{LLMs --- 5-shot}} \\
Opus 4.6        & .925 [.898,.950] & .840 [.797,.880] & .959 [.911,.991] & .852 [.803,.901] & .614 [.506,.711] & .708 [.542,.850] & .743 [.516,.903] \\
Sonnet 4.6      & .925 [.899,.950] & .812 [.765,.856] & .961 [.918,.988] & .843 [.794,.891] & .697 [.593,.777] & .600 [.428,.764] & .743 [.516,.903] \\
GPT-5.5         & .934 [.909,.955] & .810 [.767,.851] & .951 [.918,.978] & .843 [.792,.894] & .622 [.530,.696] & .550 [.412,.678] & .727 [.462,.919] \\
Opus 4.7        & .902 [.875,.926] & .803 [.758,.845] & .968 [.937,.989] & .808 [.752,.861] & .576 [.484,.664] & .514 [.338,.677] & .686 [.429,.878] \\
Gemini 2.5 Pro  & .915 [.890,.938] & .802 [.758,.842] & .904 [.848,.948] & .835 [.780,.887] & .569 [.475,.669] & .463 [.308,.595] & .688 [.400,.889] \\
Mistral Large   & .909 [.883,.933] & .790 [.745,.833] & .934 [.906,.958] & .765 [.709,.821] & .545 [.445,.638] & .432 [.277,.582] & .632 [.364,.829] \\
DeepSeek V4 Pro & .913 [.888,.936] & .772 [.721,.819] & .893 [.844,.933] & .763 [.704,.822] & .500 [.409,.604] & .466 [.312,.604] & .686 [.429,.878] \\
Grok 4.3        & .904 [.876,.930] & .691 [.636,.742] & .870 [.810,.926] & .661 [.596,.733] & .561 [.454,.667] & .384 [.219,.534] & .706 [.444,.895] \\
Llama 3.3 70B   & .874 [.836,.909] & .673 [.626,.721] & .813 [.747,.869] & .682 [.619,.743] & .577 [.482,.680] & .329 [.157,.511] & .684 [.450,.851] \\
\midrule
\multicolumn{8}{@{}l}{\textit{Fine-tuned transformers}} \\
XLM-R-large      & .947 [.926,.968] & .790 [.739,.838] & .935 [.864,.982] & .824 [.756,.880] & .584 [.474,.680] & .488 [.320,.654] & .571 [.363,.757] \\
Legal-BERT & .924 [.916,.932] & .743 [.734,.752] & .932 [.910,.954] & .816 [.792,.840] & .558 [.506,.609] & .078 [.022,.135] & .338 [.100,.577] \\
\bottomrule
\end{tabular}
\caption{Per-label F1 with 95\% document-level bootstrap confidence intervals.
Intervals widen sharply on (\textsc{soft}, \textsc{crti}):
on \textsc{crti} (18 test spans) even the strongest systems span $\sim$40 F1
points.}
\label{tab:ci-perlabel}
\end{table*}

\clearpage

\section{Agentic Loop in Action — Iteration 5}
\label{app:agentsinaction}

We reproduce verbatim excerpts from the agent communication log 
at iteration~5 of a 240-sentence pilot run ($\approx$500 
candidates), where the QA Gate identifies a distributional 
imbalance and triggers a course correction that propagates 
through the subsequent agents.

\vspace{0.5em}
\noindent\textbf{State at entry.} 74 of 240 sentences selected 
(31\%). Label counts: IL\_PROVISION=56, IL\_DOMESTIC\_LAW=33, 
IL\_TREATY=22, IL\_RESOLUTION=15, IL\_SOFT\_LAW=8, 
IL\_CASE\_CITATION=5, IL\_COURT\_INSTR=2. Corpus counts: 
ECtHR=32, UNSC=25, CD-ICJ=17.

\vspace{0.8em}
\begin{mdframed}[backgroundcolor=red!5, linecolor=red!40, 
linewidth=0.8pt, roundcorner=4pt]
\textbf{QA Gate} \hfill \texttt{pass=False}

\small\itshape
``IL\_COURT\_INSTR has only 2 selected sentences---this is the 
weakest label and needs forcing. IL\_CASE\_CITATION has only 5, 
which is also relatively low. IL\_PROVISION at 56 is heavily 
over-represented; it should be halted to allow other labels to 
catch up. Corpus ratio ECtHR/CD-ICJ = 1.88$\times$, exceeding 
the 1.5$\times$ threshold. Of 26 negatives, only 1 is hard and 
25 are empty---hard ratio is 3.8\%, far below the 50\% target. 
Swapping 12 empties for hard negatives would bring the ratio to 
$\approx$50\%.''

\vspace{0.3em}
\normalfont\small
$\rightarrow$~\texttt{halt\_labels=[IL\_PROVISION]}\\
$\rightarrow$~\texttt{force\_labels=[IL\_COURT\_INSTR,}\\
\hphantom{$\rightarrow$~}\texttt{IL\_CASE\_CITATION]}\\
$\rightarrow$~\texttt{halt\_corpora=[ECtHR]}\\
$\rightarrow$~\texttt{force\_corpora=[CD-ICJ]}\\
$\rightarrow$~\texttt{swap\_negs=12}
\end{mdframed}

\vspace{0.4em}
\begin{mdframed}[backgroundcolor=blue!5, linecolor=blue!40, 
linewidth=0.8pt, roundcorner=4pt]
\textbf{Coverage Monitor} \hfill \texttt{batch\_size=15}

\small\itshape
``The QA directive halts IL\_PROVISION (already at 56) and ECtHR 
(already leading at 32). It forces IL\_COURT\_INSTR (only 2 
selected, critically low) and IL\_CASE\_CITATION (only 5). 
CD-ICJ is forced and is the trailing corpus at 17 vs 
ECtHR's 32. Beyond the forced items, IL\_SOFT\_LAW at 8 is 
still relatively low and deserves continued attention as a 
naturally rare label.''

\vspace{0.3em}
\normalfont\small
$\rightarrow$~\texttt{priority\_labels=[IL\_COURT\_INSTR, 
IL\_CASE\_CITATION, IL\_SOFT\_LAW, IL\_RESOLUTION]}\\
$\rightarrow$~\texttt{priority\_corpora=[CD-ICJ, UNSC]}
\end{mdframed}

\newpage

\vspace{0.4em}
\begin{mdframed}[backgroundcolor=green!5, linecolor=green!40, 
linewidth=0.8pt, roundcorner=4pt]
\textbf{Selector} \hfill \texttt{+10 pos, +5 neg}

\small\itshape
Sentence \texttt{69d6}: ``Rich sentence with multiple case 
citations (ICJ reporter format, party-vs-party), a court 
instrument (Order), and a provision (Article 63).'' 
$\rightarrow$ \texttt{include=True}

\vspace{0.2em}
Sentence \texttt{898e}: ``Very rich UNSC sentence with two 
resolutions (1718, 1540), multiple provisions, two treaties 
(Chemical Weapons Convention, Biological Weapons Convention).'' 
$\rightarrow$ \texttt{include=True}

\vspace{0.2em}
Sentence \texttt{9e68}: ``Contains only organ names (UNAMSIL, 
Council, Secretary-General) and state references---none map to 
the seven-tag taxonomy.'' $\rightarrow$ 
\texttt{include=True, hard\_negative=True}
\end{mdframed}

\vspace{0.4em}
\begin{mdframed}[backgroundcolor=yellow!8, linecolor=yellow!60!black, 
linewidth=0.8pt, roundcorner=4pt]
\textbf{Diversity Gate} \hfill \texttt{kept=15, removed=0}

\small\itshape
All 15 candidates retained. 10 flagged as 
\texttt{first\_of\_kind} (introducing instruments or case names 
not yet in the comparator pool). Remaining 5 contribute along 
\texttt{argumentative\_function} (authority vs.\ distinguishing 
vs.\ listing) and \texttt{decomposition\_complexity} (composite 
citations vs.\ bare provisions).
\end{mdframed}

\vspace{0.8em}
\noindent\textbf{Outcome.} The QA Gate's directives propagate 
through subsequent iterations: the 
\texttt{halt\_labels=[IL\_PROVISION]} directive persists from 
iteration~5 through~15, during which IL\_COURT\_INSTR grows 
from 2 to 15 and IL\_CASE\_CITATION from 5 to 31. By 
iteration~20, the QA Gate issues \texttt{pass=True} with empty 
directive lists---all seven labels are represented and corpus 
balance is within threshold. The pipeline terminates after 21 
iterations with 240 selected sentences (68 LLM calls total).

\section{Few-shot Examples Assistance}
\label{app:fewshot_examples}

\begin{nerexample}{Court instrument vs. case citation}
Accordingly, an \CASE{Order of 3 September 1962} recorded that by virtue
of the provisions of \PROV{Article 62, paragraph 3}, of \CRTI{the Rules of Court},
the proceedings on the merits were suspended.

\tcblower
\footnotesize
\textit{Boundary illustrated:} a dated Order is a case-specific judicial
decision, whereas the Rules of Court are a standing procedural instrument.
\end{nerexample}

\begin{nerexample}{Treaty vs. soft law vs. resolution}
Calls upon the authorities to act in accordance with \TRTY{the ECOWAS Convention
on Small Arms and Light Weapons} and \SOFT{the United Nations Programme of Action
on Small Arms and Light Weapons}, and stresses the implementation of
\RESL{resolution 2017 (2011)}.

\tcblower
\footnotesize
\textit{Boundary illustrated:} the Convention is binding treaty law, the Programme
of Action is soft law, and the numbered resolution is an IO resolution.
\end{nerexample}

\begin{nerexample}{Composite case citation}
\CASE{Certain Activities Carried Out by Nicaragua in the Border Area
(Costa Rica v. Nicaragua), Provisional Measures, Order of 8 March 2011,
I.C.J. Reports 2011 (I), p. 18}.

\tcblower
\footnotesize
\textit{Boundary illustrated:} a full composite citation containing parties, procedural
phase, decision type, date, and reporter reference is annotated as one
case-citation span.
\end{nerexample}

\begin{nerexample}{Treaty vs. domestic law}
It follows that it is not established that her return would be in breach of
\DLAW{European Directive 2004/83/EC} or of \PROV{Article 3} of
\TRTY{the European Convention on Human Rights}.

\tcblower
\footnotesize
\textit{Boundary illustrated:} an EU Directive is secondary legislation, while
the European Convention on Human Rights is a treaty; the Article reference is
decomposed as a provision.
\end{nerexample}

\begin{nerexample}{Soft law vs. treaty vs. domestic law}
Calls upon the CAR authorities to respect their obligations under
\TRTY{the Optional Protocol to the Convention on the Rights of the Child on the
Involvement of Children in Armed Conflict}, and to consider associated children
primarily as victims as per \SOFT{the Paris Principles}, and welcomes the
adoption of \DLAW{the child protection code}.

\tcblower
\footnotesize
\textit{Boundary illustrated:} a ratified protocol is treaty law, endorsed
principles are soft law, and an adopted code is domestic law.
\end{nerexample}

\begin{nerexample}{Resolution symbols and soft law}
Concerned that the parties' continuing suspicion and lack of trust have
contributed to delays in the implementation of \SOFT{the Settlement Plan}
(\RESL{S/21360} and \RESL{S/22464} and Corr.1).

\tcblower
\footnotesize
\textit{Boundary illustrated:} UN document symbols are resolutions/documents,
whereas a named plan without binding treaty status is soft law.
\end{nerexample}

\begin{nerexample}{EU regulations and provision decomposition}
This resolution was implemented by \DLAW{Regulation (EC) no. 2472/94} of
10 October 1994, \PROV{Article 5} of which suspended the operation of
\PROV{Article 8} of \DLAW{Regulation (EEC) no. 990/93}.

\tcblower
\footnotesize
\textit{Boundary illustrated:} EU Regulations are domestic/supranational
secondary legislation, while their Articles are separate provision spans.
\end{nerexample}

\begin{nerexample}{Numbered resolution with case citation}
General Assembly \RESL{resolution 2625 (XXV)} mentioned in
\PROV{paragraph 162} of \CASE{the Judgment} does not, in my view, apply to
the present case.

\tcblower
\footnotesize
\textit{Boundary illustrated:} a numbered General Assembly resolution is tagged
as a resolution; the Judgment is a case citation; the paragraph reference is a
locator/provision-style reference.
\end{nerexample}

\section{The Full Label Glossary \& Annotation Guidelines}
\label{app:definitions}

This appendix reports the full label glossary used in the annotation
guidelines and LLM prompts. 
The types were fixed deductively from the sources of international law (ICJ Statute Art.~38(1)) and the IO-law literature on resolutions and soft law, then pruned inductively to categories that surface as discrete named entities in the three corpora at annotatable frequency.

The schema contains seven canonical labels.
No other labels are valid.

\subsection{Canonical Labels}

\paragraph{\texttt{IL\_TREATY}: Binding international agreements.}
An international agreement concluded between states or international
organizations in written form and governed by international law,
regardless of its particular designation \citep[Art.~2(1)(a)]{vclt1969}
The test is whether the instrument creates binding legal obligations, not
what it is called.

\textbf{Includes:}
\begin{itemize}[leftmargin=*,topsep=2pt,itemsep=1pt]
    \item Multilateral conventions, covenants, charters, protocols, and pacts.
    \item Constitutive treaties of international organizations.
    \item Bilateral treaties and Bilateral Investment Treaties (BITs).
    \item Peace agreements, ceasefire agreements, and boundary agreements that create binding obligations between states, even when titled ``Declaration'' or ``Accord''.
    \item Statutes of international courts and tribunals when annexed to or adopted under treaty authority.
    \item Acronyms standing for treaty instruments, such as ECHR, ICCPR, ICESCR, CEDAW, CERD, CAT, CRC, CRPD, CISG, UNCAT, ACHR, ACHPR, and VCLT.
\end{itemize}

\textbf{Recognition patterns:}
\begin{itemize}[leftmargin=*,topsep=2pt,itemsep=1pt]
    \item ``Convention on/against/for [topic]'' or ``[topic] Convention''.
    \item Year plus treaty name, e.g., ``1949 Geneva Conventions''.
    \item ``Protocol No.~N'' when naming an entire additional protocol as a standalone instrument.
    \item ``Statute of the [Court/Tribunal]''.
    \item Anaphoric references such as ``the Convention'', ``the Charter'', or ``the Statute'' when there is a clear antecedent.
\end{itemize}

\textbf{Canonical examples:}
\emph{the Charter of the United Nations}; \emph{the UN Charter};
\emph{the Vienna Convention on the Law of Treaties}; \emph{the
International Covenant on Civil and Political Rights}; \emph{the ICCPR};
\emph{the European Convention on Human Rights}; \emph{Protocol No.~1 to
the Convention}; \emph{Protocol No.~12}; \emph{the Rome Statute};
\emph{the Statute of the International Court of Justice}; \emph{the
Geneva Conventions of 12 August 1949}; \emph{Additional Protocol I};
\emph{the Argentina--United States BIT}; \emph{the Treaty of Amity,
Economic Relations, and Consular Rights}; \emph{the Dayton Peace
Agreement}; \emph{the Maroua Declaration}; \emph{the Comprehensive Peace
Agreement}.

\textbf{Does not include:}
\begin{itemize}[leftmargin=*,topsep=2pt,itemsep=1pt]
    \item Rules of Court, Rules of Procedure, and Practice Directions, which are \texttt{IL\_COURT\_INSTR}.
    \item EU Regulations and Directives, which are \texttt{IL\_DOMESTIC\_LAW}.
    \item Non-binding declarations and guidelines, which are \texttt{IL\_SOFT\_LAW}.
\end{itemize}

\paragraph{\texttt{IL\_PROVISION}: Numbered subdivisions of legal instruments.}
Internal subdivisions of a treaty, resolution, statute, domestic law,
rules of procedure, or other legal instrument. These include articles,
paragraphs, sections, clauses, chapters, parts, rules, subparagraphs,
annexes, and schedules.

\textbf{Recognition patterns:}
\begin{itemize}[leftmargin=*,topsep=2pt,itemsep=1pt]
    \item ``Article N'', ``Article N(M)'', ``Article N, paragraph M'', or ``Art.~N''.
    \item ``paragraph N'', ``paragraphs N and M'', ``para.~N'', ``\S~N'', or ``\S\S~N--M''.
    \item ``section N'', ``Section II'', or ``section 65(3)''.
    \item ``Chapter VII'' or ``Chapter III of the Charter''.
    \item ``Rule N'', ``Rule 39'', or ``Rule 22 bis''.
    \item ``subparagraph (a)'' or ``clause 2''.
    \item ``Annex VII'', ``Annex 1-A'', ``the First Schedule'', or ``Second Schedule''.
    \item ``Part III'', ``Title X'', or ``Code 3.1''.
\end{itemize}

\textbf{Canonical examples:}
\emph{Article 2(4)}; \emph{Article 6}; \emph{Article 5 \S~1(c)};
\emph{Article 38(1) of the Statute}; \emph{Article 36, paragraph 2};
\emph{paragraph 24}; \emph{para.~138}; \emph{\S~14}; \emph{\S\S~32--45};
\emph{Chapter VII}; \emph{Rule 39}; \emph{Rule 65 bis};
\emph{section 65(3)}; \emph{Annex VII}; \emph{Annex 1-A};
\emph{the First Schedule}; \emph{Code 3.1}.

\textbf{Does not include:}
\begin{itemize}[leftmargin=*,topsep=2pt,itemsep=1pt]
    \item ``Protocol No.~N'' when naming an entire protocol as an instrument; this is \texttt{IL\_TREATY}.
    \item Discourse references such as ``the preceding paragraph'' or ``the paragraph above''.
    \item Paragraph references within judicial decisions when they are part of a case citation; these belong inside the \texttt{IL\_CASE\_CITATION} span.
    \item Pinpoint locators within judgments, e.g., ``para.~32'' in ``the Judgment at paragraph 32'', which are treated as part of \texttt{IL\_CASE\_CITATION}, not as standalone provisions.
\end{itemize}

\paragraph{\texttt{IL\_RESOLUTION}: Formally numbered decisions of UN or IO organs.}
A formally numbered decision, recommendation, or document of an
international organization. UN document symbols are high-precision
signals for this label. The label is assigned irrespective of binding force: Security Council and General Assembly resolutions are both IL\_RESOLUTION when cited by number or symbol.

\textbf{UN document-symbol patterns:}
\begin{itemize}[leftmargin=*,topsep=2pt,itemsep=1pt]
    \item \texttt{S/RES/\#\#\#\#}: Security Council resolution.
    \item \texttt{A/RES/\#\#/\#\#}: General Assembly resolution.
    \item \texttt{A/HRC/RES/\#\#/\#\#}: Human Rights Council resolution.
    \item \texttt{S/PRST/\#\#\#\#/\#\#}: Security Council Presidential Statement.
    \item \texttt{S/\#\#\#\#/\#\#\#\#}: Security Council document.
    \item \texttt{A/\#\#/\#\#\#}: General Assembly document.
    \item \texttt{E/RES/\#\#\#\#/\#\#}: ECOSOC resolution.
\end{itemize}

\textbf{Surface-form patterns:}
\begin{itemize}[leftmargin=*,topsep=2pt,itemsep=1pt]
    \item ``resolution N (YEAR)'', e.g., ``resolution 1970 (2011)''.
    \item ``resolution N/M of DATE''.
    \item ``Security Council resolution N (YEAR)''.
    \item ``General Assembly resolution N/M''.
    \item ``EU Council Decision YEAR/NNN/CFSP''.
\end{itemize}

\textbf{Canonical examples:}
\emph{resolution 1325 (2000)}; \emph{resolution 1970 (2011)};
\emph{resolution 2589}; \emph{S/RES/2589}; \emph{A/RES/70/1};
\emph{A/HRC/RES/29/11}; \emph{Security Council resolution 1373 (2001)};
\emph{A/45/594}; \emph{S/1996/1012}.

\textbf{Does not include:}
\begin{itemize}[leftmargin=*,topsep=2pt,itemsep=1pt]
    \item Descriptive uses of ``resolution'' without a formal number.
    \item UNGA instruments cited by declaration name rather than by number, e.g., \emph{the Universal Declaration of Human Rights}, which is \texttt{IL\_SOFT\_LAW}.
\end{itemize}

\paragraph{\texttt{IL\_COURT\_INSTR}: Standing procedural instruments of courts and tribunals.}
Procedural instruments adopted by a court or tribunal under its own
authority to regulate its operations on a continuing basis. These are not
case-specific decisions. The critical distinction is that a court
instrument is a persistent procedural document governing the institution,
whereas a judgment, order, advisory opinion, or award is a decision in a
specific case.

\textbf{Includes:}
\begin{itemize}[leftmargin=*,topsep=2pt,itemsep=1pt]
    \item Rules of Court.
    \item Rules of Procedure and Evidence.
    \item Practice Directions.
    \item Provisional Rules of Procedure.
    \item Resolutions concerning internal judicial practice.
    \item Regulations of the Court.
\end{itemize}

\textbf{Recognition patterns:}
\begin{itemize}[leftmargin=*,topsep=2pt,itemsep=1pt]
    \item ``the Rules of Court'' or ``Rules of the Court''.
    \item ``the Rules of Procedure'' or ``Rules of Procedure and Evidence''.
    \item ``Practice Direction [N/Roman]''.
    \item ``the provisional rules of procedure''.
    \item ``Regulations of the Court''.
\end{itemize}

\textbf{Canonical examples:}
\emph{the Rules of Court}; \emph{the Rules of Procedure and Evidence};
\emph{Practice Direction V}; \emph{the provisional rules of procedure of
the Security Council}; \emph{the Regulations of the Court}.

\textbf{Does not include:}
\begin{itemize}[leftmargin=*,topsep=2pt,itemsep=1pt]
    \item Judgments, Orders, and Advisory Opinions, which are \texttt{IL\_CASE\_CITATION}.
    \item Arbitral awards, which are \texttt{IL\_CASE\_CITATION}.
    \item Memorials, Counter-Memorials, Applications, and Observations, which are party submissions and receive no tag under the current schema.
    \item Statutes of courts and tribunals adopted under treaty authority, which are \texttt{IL\_TREATY}.
    \item Sanctions lists and committee reports, which receive no tag.
\end{itemize}

\paragraph{\texttt{IL\_CASE\_CITATION}: References to specific judicial or arbitral decisions.}
Any reference to a specific judicial or arbitral decision, whether by
case name, party names, decision type, reporter citation, or docket
number. This label covers the cited decision as a legal authority.

\textbf{Includes:}
\begin{itemize}[leftmargin=*,topsep=2pt,itemsep=1pt]
    \item Party-vs-party case names, e.g., ``X v. Y''.
    \item Named cases without party names, e.g., ``the Corfu Channel case''.
    \item Judicial decision types when referring to a specific case: judgments, orders, advisory opinions, arbitral awards, and decisions on jurisdiction, admissibility, merits, or annulment.
    \item Reporter citations, such as I.C.J. Reports, P.C.I.J. Series, and ECHR Series A.
    \item Application or docket numbers.
    \item Composite citations combining case names, decision types, reporter references, and pinpoint references.
    \item Cross-references such as ``ibid.'', ``supra'', and ``op. cit.'' when they point to a specific decision.
\end{itemize}

\textbf{Recognition patterns:}
\begin{itemize}[leftmargin=*,topsep=2pt,itemsep=1pt]
    \item ``X v. Y'' or ``X c. Y''.
    \item ``Judgment of [DATE]'', ``Order of [DATE]'', or ``Advisory Opinion of/on [topic]'' when referring to a specific decision.
    \item Anaphoric references such as ``the Judgment'', ``the Order'', ``the Advisory Opinion'', or ``the present Order'' when they refer to a specific decision.
    \item ``I.C.J. Reports YEAR, p.~N'' or ``P.C.I.J. Series A''.
    \item ``Application no.~N/YEAR'' or ``no.~N/YEAR''.
    \item ``Case No.~IT-\#\#-\#\#'', ``ICC-\#\#/\#\#-\#\#/\#\#'', or ``ICSID Case No.~ARB/\#\#/\#\#''.
    \item ``the [Name] case''.
    \item ``the Cleveland Award'' or ``the arbitral award of [DATE]''.
    \item ``Judgement No.~N of the [Tribunal]''.
    \item French forms such as ``ordonnance du [DATE]'' or ``arr\^et du [DATE]''.
\end{itemize}

\textbf{Composite citation rule.}
When a case name, decision type, reporter citation, and pinpoint
paragraph appear together as one citation, the entire citation is tagged
as a single \texttt{IL\_CASE\_CITATION} span. For example:
\begin{itemize}[leftmargin=*,topsep=2pt,itemsep=1pt]
    \item \emph{Nicaragua v. United States, Merits, Judgment, I.C.J. Reports 1986, p.~14, para.~178}.
    \item \emph{Soering v. the United Kingdom, 7 July 1989, \S~88, Series A no.~161}.
    \item \emph{Application of the Convention on the Prevention and Punishment of the Crime of Genocide (Bosnia and Herzegovina v. Serbia and Montenegro), Judgment, I.C.J. Reports 2007, p.~43}.
\end{itemize}

Some ICJ case names embed treaty names, for example
\emph{Application of the Genocide Convention (Croatia v. Serbia)}. When
the treaty name is part of the official case name, the entire case name
is tagged as \texttt{IL\_CASE\_CITATION}. If the same treaty appears
independently elsewhere in the sentence, it is tagged as
\texttt{IL\_TREATY}.

\textbf{Canonical examples:}
\emph{Nicaragua v. United States}; \emph{Bosnia and Herzegovina v.
Serbia and Montenegro}; \emph{the Corfu Channel case}; \emph{Nottebohm};
\emph{LaGrand}; \emph{I.C.J. Reports 1986, p.~14}; \emph{Order of 10
July 2002}; \emph{the Judgment}; \emph{Advisory Opinion on Nuclear
Weapons}; \emph{the Advisory Opinion}; \emph{Application no.~29731/96};
\emph{the Cleveland Award}; \emph{ordonnance du 2 mai 2019};
\emph{Judgement No.~158 of the United Nations Administrative Tribunal};
\emph{CMS Gas Transmission v. Argentina, ICSID Case No.~ARB/01/8};
\emph{Klass v. Germany, no.~5029/71}.

\textbf{Does not include:}
\begin{itemize}[leftmargin=*,topsep=2pt,itemsep=1pt]
    \item Self-references such as ``the present case'' or ``the instant case''.
    \item Rules of Court and Practice Directions, which are \texttt{IL\_COURT\_INSTR}.
    \item Party submissions such as Memorials and Counter-Memorials, which receive no tag.
\end{itemize}

\paragraph{\texttt{IL\_SOFT\_LAW}: Non-binding international instruments.}
Declarations, communiqu\'es, action plans, guidelines, principles, codes
of conduct, frameworks, and agendas that do not create enforceable legal
obligations between states.

The test is whether the instrument was concluded as a binding agreement
between states governed by international law. Surface form alone is not
decisive: instruments titled ``Declaration'' or ``Agreement'' may still
be \texttt{IL\_TREATY} if they create binding obligations. Conversely,
the Universal Declaration of Human Rights is \texttt{IL\_SOFT\_LAW}
because it was adopted as a UN General Assembly declaration, even though
parts of it may reflect customary international law.

\textbf{Recognition patterns:}
\begin{itemize}[leftmargin=*,topsep=2pt,itemsep=1pt]
    \item ``[Place/Topic] Declaration'' or ``Declaration on [topic]'', after verifying non-binding status.
    \item ``[Topic] Communiqu\'e'' or ``Joint Communiqu\'e''.
    \item ``Plan of Action'' or ``[Year] Action Plan''.
    \item ``Guiding Principles on [topic]'' or ``Basic Principles on [topic]''.
    \item ``[Place/Year] Framework'', such as Sendai or Hyogo.
    \item ``[Year] Agenda'', such as the 2030 Agenda.
    \item ``Code of Conduct for [actor]''.
\end{itemize}

\textbf{Canonical examples:}
\emph{the Universal Declaration of Human Rights}; \emph{the UDHR};
\emph{the Rio Declaration}; \emph{the Stockholm Declaration};
\emph{the Vienna Declaration and Programme of Action}; \emph{the Safe
Schools Declaration}; \emph{the Geneva Communiqu\'e};
\emph{the Guiding Principles on Internal Displacement};
\emph{the Sendai Framework}; \emph{the 2030 Agenda};
\emph{the Code of Conduct for Law Enforcement Officials}.

\textbf{Does not include:}
\begin{itemize}[leftmargin=*,topsep=2pt,itemsep=1pt]
    \item Binding covenants such as the ICCPR and ICESCR, which are \texttt{IL\_TREATY}.
    \item Binding peace or boundary agreements titled ``Declaration'', which are \texttt{IL\_TREATY}.
    \item UNGA resolutions cited by number, which are \texttt{IL\_RESOLUTION}.
    \item Unilateral state declarations, which receive no tag.
\end{itemize}

\paragraph{\texttt{IL\_DOMESTIC\_LAW}: National and supranational secondary legislation.}
National statutes, constitutions, codes, decrees, and secondary
legislation cited in international proceedings. This label also includes secondary legislation of 
supranational bodies such as EU Regulations and Directives, 
which are adopted through internal legislative procedures rather 
than negotiated as inter-state treaties.

\textbf{Includes:}
\begin{itemize}[leftmargin=*,topsep=2pt,itemsep=1pt]
    \item National statutes, codes, acts, and organic laws.
    \item Constitutions and constitutional amendments.
    \item Decrees, decree-laws, presidential decrees, and royal decrees.
    \item Orders in Council and Statutory Instruments.
    \item EU Regulations and EU Directives, as secondary legislation rather than treaties.
\end{itemize}

\textbf{Recognition patterns:}
\begin{itemize}[leftmargin=*,topsep=2pt,itemsep=1pt]
    \item ``[Topic] Code'', ``Criminal Code'', or ``Civil Code''.
    \item ``[Topic] Act [year]'' or ``[Topic] Act of [year]''.
    \item ``the Constitution'' or ``the Constitution of [State]''.
    \item ``[Ordinal] Amendment''.
    \item ``Law no.~N/YEAR'' or ``Law No.~N of YEAR''.
    \item ``Decree-Law No.~N'' or ``Presidential Decree No.~N''.
    \item ``Council Regulation (EC) No.~N/YEAR'' or ``Directive YEAR/N/EC''.
    \item ``Order in Council'' or ``Statutory Instrument no.~N''.
\end{itemize}

\textbf{Canonical examples:}
\emph{the Criminal Code}; \emph{the Code of Criminal Procedure};
\emph{the Human Rights Act 1998}; \emph{the Constitution};
\emph{the First Amendment}; \emph{Law no.~229/1991};
\emph{Presidential Decree No.~5}; \emph{Council Regulation (EC)
No.~343/2003}; \emph{the Brussels II bis Regulation};
\emph{the British Protectorates, Protected States and Protected Persons
Order in Council, 1949}; \emph{Statutory Instrument no.~144 of 1993}.

\textbf{Does not include:}
\begin{itemize}[leftmargin=*,topsep=2pt,itemsep=1pt]
    \item Descriptors such as ``domestic law'' or ``national legislation''.
    \item EU founding treaties, such as the TEU and TFEU, which are \texttt{IL\_TREATY}.
\end{itemize}

\subsection{General Exclusions}

The following are never tagged as entities under the schema:
\begin{itemize}[leftmargin=*,topsep=2pt,itemsep=1pt]
    \item Organ or institution names, such as ``Security Council'', ``the Court'', ``the General Assembly'', ``the Trial Chamber'', or ``the Registry''.
    \item State or party names, such as ``France'', ``the Respondent'', or ``the Applicant''.
    \item Discourse anaphora, such as ``the present case'', ``this resolution'', or ``the said Convention''.
    \item Generic legal descriptors, such as ``domestic law'', ``international law'', ``national legislation'', or ``customary international law''.
    \item Party pleadings and submissions, such as ``the Memorial of Nicaragua'', ``the Counter-Memorial'', ``Application instituting proceedings'', ``the Preliminary Objections'', and ``Written Observations''.
    \item Administrative documents of subsidiary bodies, such as sanctions lists, committee reports, and consolidated lists.
\end{itemize}

Foreign-language equivalents follow the same rules. For example,
\emph{Convention europ\'eenne des droits de l'homme} is
\texttt{IL\_TREATY}; \emph{C\'odigo Penal} is
\texttt{IL\_DOMESTIC\_LAW}; \emph{Malaisie c. Singapour} is
\texttt{IL\_CASE\_CITATION}; and \emph{ordonnance du 2 mai 2019} is
\texttt{IL\_CASE\_CITATION}.

\subsection{Decomposition Rules}

Composite legal references are decomposed into atomic, non-overlapping
spans unless the entire string functions as a single case citation.

\begin{itemize}[leftmargin=*,topsep=2pt,itemsep=2pt]
    \item ``Article 2(4) of the UN Charter'' is two entities:
    \begin{itemize}[topsep=1pt,itemsep=1pt]
        \item \emph{Article 2(4)} $\rightarrow$ \texttt{IL\_PROVISION}
        \item \emph{the UN Charter} $\rightarrow$ \texttt{IL\_TREATY}
    \end{itemize}

    \item ``paragraph 24 of resolution 1970 (2011)'' is two entities:
    \begin{itemize}[topsep=1pt,itemsep=1pt]
        \item \emph{paragraph 24} $\rightarrow$ \texttt{IL\_PROVISION}
        \item \emph{resolution 1970 (2011)} $\rightarrow$ \texttt{IL\_RESOLUTION}
    \end{itemize}

    \item ``Article 38(1) of the Statute of the Court'' is two entities:
    \begin{itemize}[topsep=1pt,itemsep=1pt]
        \item \emph{Article 38(1)} $\rightarrow$ \texttt{IL\_PROVISION}
        \item \emph{the Statute of the Court} $\rightarrow$ \texttt{IL\_TREATY}
    \end{itemize}

    \item ``Rule 39 of the Rules of Court'' is two entities:
    \begin{itemize}[topsep=1pt,itemsep=1pt]
        \item \emph{Rule 39} $\rightarrow$ \texttt{IL\_PROVISION}
        \item \emph{the Rules of Court} $\rightarrow$ \texttt{IL\_COURT\_INSTR}
    \end{itemize}
\end{itemize}

Case citations are the main exception. When the whole string is a case
identifier, the entire citation remains one \texttt{IL\_CASE\_CITATION}
span:
\begin{itemize}[leftmargin=*,topsep=2pt,itemsep=1pt]
    \item \emph{Nicaragua v. United States, Merits, Judgment, I.C.J. Reports 1986, p.~14}
    $\rightarrow$ one \texttt{IL\_CASE\_CITATION}.
    \item \emph{Soering v. the United Kingdom, 7 July 1989, \S~88, Series A no.~161}
    $\rightarrow$ one \texttt{IL\_CASE\_CITATION}.
\end{itemize}

\end{document}